\documentclass[letterpaper, 10 pt, conference]{ieeeconf}
\IEEEoverridecommandlockouts
\usepackage{graphics} 
\usepackage{epsfig} 
\usepackage{times} 
\usepackage{amsmath} 
\usepackage{amssymb}  
\usepackage{url}
\usepackage{algpseudocode}
\usepackage{algorithm}
\usepackage{algorithmicx}
\usepackage{multirow}
\usepackage{xcolor}
\usepackage{makecell}
\usepackage{threeparttable}
\usepackage{wrapfig}
\makeatletter
\let\NAT@parse\undefined
\makeatother
\usepackage[colorlinks=true, citecolor=blue, linkcolor=blue, urlcolor=blue]{hyperref}

\DeclareMathOperator*{\argmax}{arg\,max}

\DeclareMathOperator*{\eventually}{\Diamond}
\DeclareMathOperator*{\always}{\Box}
\DeclareMathAlphabet{\mathmybb}{U}{bbold}{m}{n}

\title{\LARGE \bf
  STL-SVPIO: Signal Temporal Logic guided Stein Variational \\Path Integral Optimization
}

\author{Hongrui Zheng$^{1}$, Zirui Zang$^{1}$, Ahmad Amine$^{1}$, Cristian Ioan Vasile$^{2}$, Rahul Mangharam$^{1}$%
\thanks{$^{1}$Department of Electrical and Systems Engineering, University of Pennsylvania, Philadelphia, PA, USA. $^{2}$Mechanical Engineering and Mechanics Department, Lehigh University, Bethlehem, PA, USA. Correspond to \tt\small hongruiz@seas.upenn.edu.}%
}

\begin{document}

\maketitle
\thispagestyle{empty}
\pagestyle{empty}

\begin{abstract}
Signal Temporal Logic (STL) enables formal specification of complex spatiotemporal constraints for robotic task planning. However, synthesizing long-horizon continuous control trajectories from complex  STL specifications is fundamentally challenging due to the nested structure of STL robustness objectives. Existing solver-based methods, such as Mixed-Integer Linear Programming (MILP), suffer from exponential scaling, whereas sampling methods, such as Model-Predictive Path Integral control (MPPI), struggle with sparse, long-horizon costs. We introduce Signal Temporal Logic guided Stein Variational Path Integral Optimization (STL-SVPIO), which reframes STL as a globally informative, differentiable reward-shaping mechanism. By leveraging Stein Variational Gradient Descent and differentiable physics engines, STL-SVPIO transports a mutually repulsive swarm of control particles toward high robustness regions. Our method transforms sparse logical satisfaction into tractable variational inference, mitigating the severe local minima traps of standard gradient-based methods. We demonstrate that STL-SVPIO significantly outperforms existing methods in both robustness and efficiency for traditional STL tasks. Moreover, it solves complex long-horizon tasks, including multi-agent coordination with synchronization and queuing while baselines either fail to discover feasible solutions, or become computationally intractable. Finally, we use STL-SVPIO in agile robotic motion planning tasks with nonlinear dynamics, such as 7-DoF manipulation and half cheetah back flips to show the generalizability of our algorithm.
\end{abstract}
\section{Introduction}
Modern robotic and autonomous systems, including multi-agent teams, high-dimensional manipulators, and humanoids, are increasingly required to execute temporally extended and complex behaviors. In reinforcement learning, these objectives are typically encoded through meticulously designed reward functions~\cite{sutton_reinforcement_2018}.
However, for tasks with rich temporal structures, rewards are often challenging to engineer, resulting in sparse and delayed feedback that exacerbates temporal credit assignment and yields brittle solutions~\cite{sutton_reinforcement_2018,barto2003recent}. Although recent advances in reward design via LLMs~\cite{ma2024eureka} show impressive results in dexterous manipulation and locomotion tasks, their applicability to long-horizon planning problems with explicit temporal-logic constraints is currently unclear. On the other hand, Signal Temporal Logic (STL) provides a mathematically rigorous and expressive specification language for complex spatiotemporal task requirements in continuous systems~\cite{fainekos2009robustness,donze2010robust}. In contrast to reward-based formulations, positive STL robustness directly certifies task satisfaction. Despite its expressiveness, synthesizing continuous high-dimensional control trajectories that satisfy STL specifications remains a notoriously difficult open problem. As the planning horizon increases, the feasible solution set induced by conjunction-heavy, deadline-constrained STL often shrinks drastically, resulting in severe optimization difficulties ~\cite{donze2010robust,raman_model_2014}. Even when the problem is feasible, the size of the optimization problem in traditional STL solvers (e.g. MILP) grows exponentially with specification complexity in the worst case, often rendering them computationally intractable for realistic robotic systems~\cite{raman_model_2014,kurtz2022mixed}.

The overarching goal of this paper is to enable tractable, continuous control for highly complex spatiotemporal tasks by synergizing SVGD and STL: leveraging STL for expressive, differentiable reward shaping, while utilizing SVGD to navigate the multi-modal optimization landscapes. Our contributions in this paper are threefold:
\begin{enumerate}
    \item We introduce STL-SVPIO, a variational inference framework for long-horizon task and motion planning. We provide a principled derivation of the resulting inference formulation.
    \item We show that STL specifications can be effectively used as a differentiable shaping signal for motion planning, extending their applicability to agile robotic behaviors without relying on handcrafted rewards.
    \item We analyze how particle-based variational inference mitigates the non-convexity and multi-modality in cost induced by STL constraints, and demonstrate our method in task satisfaction.
\end{enumerate}
We showcase STL-SVPIO across a diverse set of tasks, including long-horizon navigation, multi-agent coordination with synchronization and ordering constraints, agile locomotion, and torque-controlled manipulation. The paper is organized as follows: in Section \ref{sec:formulation}, we introduce our variational inference problem; in Section \ref{sec:method}, we introduce our proposed framework and how it modifies the original inference problem; and in Section \ref{sec:exp}, we present simulation results on various benchmarks.

\section{Related Work}

STL provides a quantitative robustness evaluation that indicates how strongly the specification is satisfied or violated \cite{fainekos2009robustness,donze2010robust}. Robot task and motion planning under STL specifications remains challenging. The existing literature on control synthesis under STL specifications can be broadly organized into three categories:

\textit{Mixed Integer Linear Programming} (MILP) yields exact solutions to STL specifications under chosen discretization and relaxations~\cite{belta_formal_2019,raman_model_2014}, and has been widely applied to Model Predictive Control (MPC) \cite{sadraddini_formal_2019} and multi-agent motion planning \cite{sun_multi-agent_2022, buyukkocak2021planning}. Although MILP solvers provide clean solutions to simple problems, they scale poorly with specification complexity. While improved encodings \cite{kurtz2022mixed} and partial-satisfaction frameworks \cite{cardona_partial_2023} mitigate this issue, MILP becomes intractable for highly non-linear systems and complicated collaborative tasks~\cite{belta_formal_2019}.

\textit{Gradient-based Optimization methods} have been proposed to alleviate the computational burden associated with MILP-based solvers.
Smooth approximations are proposed to enable gradient descent for control synthesis \cite{pant_smooth_2017,mehdipour2019arithmetic}. This concept was extended by STLCG \cite{leung2023backpropagation,kapoor_stlcg_2025}, which parses STL formulae into computation graphs and obtains the gradients via backpropagation. This facilitated the use of neural networks in STL planning \cite{meng2023signal}. However, they are susceptible to being trapped in local optima due to the non-convex landscape of STL robustness, severely limiting their practicality.

\textit{Sampling-based Methods} provide a probabilistic alternative for handling the non-convexity of STL specifications. Rapidly-Exploring Random Trees (RRT) are used to search for feasible trajectories under STL specifications~\cite{vasile2017sampling,verhagen2024temporally,marchesini_sampling-based_2025}. These approaches either use STL fragments or impose strict constraints on the admissible sampling domain. Although effective in simple task requirements, these methods are not directly comparable to our proposed approach. Building on path integral control theory \cite{kappen_path_2005,williams_aggressive_2016}, deterministic path integral optimization is used to directly minimize STL costs \cite{halder_trajectory_2025}. However, these sampling-based methods are fundamentally bottlenecked by the curse of dimensionality. As the planning horizon and task complexity increase, the probability of randomly sampling a valid sequence decays exponentially, making convergence increasingly unlikely in practice. 

Our work leverages \textit{Stein Variational Gradient Descent} (SVGD)~\cite{liu_stein_2016}. Unlike standard gradient descent, which can collapse to a single mode, SVGD maintains a diverse set of particles that approximate the target distribution. This method has been utilized in Stein Variational Model Predictive Control (SVMPC) \cite{lambert_stein_2021}, which offers improved exploration capabilities compared to Model Predictive Path Integral (MPPI) control \cite{williams_aggressive_2016}.
To the best of our knowledge, STL-SVPIO is the first framework to explicitly bridge differentiable STL with Stein Variational inference, using the logic's gradient as a global guiding potential to solve the long-horizon sparse problem that poses significant challenges for existing approaches.

\section{Problem Formulation}\label{sec:formulation}

\subsection{Preliminary: Signal Temporal Logic (STL)}\label{sec:background_stl}
\textbf{Syntax and Grammar:}
STL expresses spatiotemporal requirements over real-valued trajectories $s$~\cite{maler_monitoring_2004}. In this work, we use STL formulae to describe the tasks assigned to agents.
The syntax of STL is defined as
\begin{equation}\label{eq:stl_syntax}
  \phi := \top \mid \mu^g \mid \neg\phi \mid \phi_1\wedge \phi_2 \mid \phi_1\,\mathbf{U}_{[t_1,t_2]}\,\phi_2,
\end{equation}
where $\top$ is the truth value, $\mu^g:=g(s(t))\ge 0$ is an atomic predicate whose truth value is determined by the value of $g:\mathbb{R}^n\rightarrow\mathbb{R}$.
Symbols $\wedge$ and $\neg$ are Boolean operators ``conjunction" and ``negation" respectively.
$\mathbf{U}$ is the temporal operator ``until", which denotes that formula $\phi_1$ has to hold until $\phi_2$ holds in the time interval $[t_1,t_2]$.
Other common operators can be defined in terms of these primitives.
Disjunction can be defined as $\phi_1 \vee \phi_2 := \neg(\neg\phi_1 \wedge \neg\phi_2)$.
``Eventually" can be defined as $\eventually_{[t_1,t_2]}\phi := \top\,\mathbf{U}_{[t_1,t_2]}\,\phi$.
``Always" can be defined as $\always_{[t_1,t_2]}\phi := \neg\eventually_{[t_1,t_2]}\neg\phi$. 
``Implication" can be defined as $\phi_1\implies\phi_2:=\neg\phi_1\vee\phi_2$.

\textbf{STL Quantitative Semantics:}
To measure how strongly a signal satisfies or violates $\phi$, STL admits a quantitative semantics called \textbf{robustness} $\rho^\phi(s,t)\in\mathbb{R}$ such that $\rho>0$ implies satisfaction and $\rho<0$ implies violation~\cite{donze2010robust,fainekos2009robustness}. For the syntax in Equation \eqref{eq:stl_syntax}, a standard robustness on a signal $s$ is defined as follows:
\begin{equation}\label{eq:robustness_defn}
  \begin{aligned}
    \rho^\top&=\infty, \\
    \rho^{\mu^g}(s,t) &= g(s(t)), \\
    \rho^{\neg\phi}(s,t) &= -\rho^{\phi}(s,t),\\
    \rho^{\phi_1\wedge\phi_2}(s,t) &= \min\big(\rho^{\phi_1}(s,t),\rho^{\phi_2}(s,t)\big),\\
    \rho^{\phi_1\,\mathbf{U}_{[a,b]}\,\phi_2}(s,t) &= \max_{t'\in[t+a,t+b]} \min\!\Big(\rho^{\phi_2}(s,t'),\\
    &\min_{t''\in[t,t']} \rho^{\phi_1}(s,t'')\Big).
  \end{aligned}
\end{equation}

\textbf{Differentiable STL:}
Recent work~\cite{pant_smooth_2017,leung2023backpropagation,kapoor_stlcg_2025} introduced smooth max and min approximators (e.g., LogSumExp) and subsequently automatic parsing of STL into computational graphs. This enables backpropagation through the robustness value of a trajectory to the states in the trajectory. With differentiable dynamics, this enables end-to-end gradient computation.

\subsection{Optimization Problem}

Consider a time-invariant discrete-time dynamical system $\mathbf{x}_{t+1} = f(\mathbf{x}_t, \mathbf{u}_t)$ where $\mathbf{x} \in \mathcal{X}$ is the state and $\mathbf{u} \in \mathcal{U}$ is the control input. Bold variables denote vectors.
We obtain the state trajectory by integrating the dynamics: $\mathbf{x}_{0:H}=\{\mathbf{x}_0, f(\mathbf{x}_0,\mathbf{u}_0), f(f(\mathbf{x}_0,\mathbf{u}_0), \mathbf{u}_1), \ldots\}$.
We define our optimization objective as the robustness value $\rho^\phi(\mathbf{x}_{0:H})$, which quantifies how well a trajectory satisfies $\phi$.
For a given finite-horizon $H$, we seek a control sequence $\mathbf{u} = \{\mathbf{u}_0, \dots, \mathbf{u}_{H-1}\}$ that satisfies a task specification $\phi$ defined in STL by maximizing the robustness score:

\begin{equation}
\begin{aligned}
    \mathbf{u}_{0:H-1}^*&=\argmax_{\mathbf{u}\in\mathcal{U}}(\rho^\phi(\mathbf{x}_{0:H})), \\
    \text{s.t. }~~ \mathbf{x}_{t+1}&=f(\mathbf{x}_t,\mathbf{u}_t) \quad \forall t \in [0 \dots H-1].
\end{aligned}
\end{equation}

\begin{figure*}
    \centering
    \includegraphics[width=0.95\linewidth]{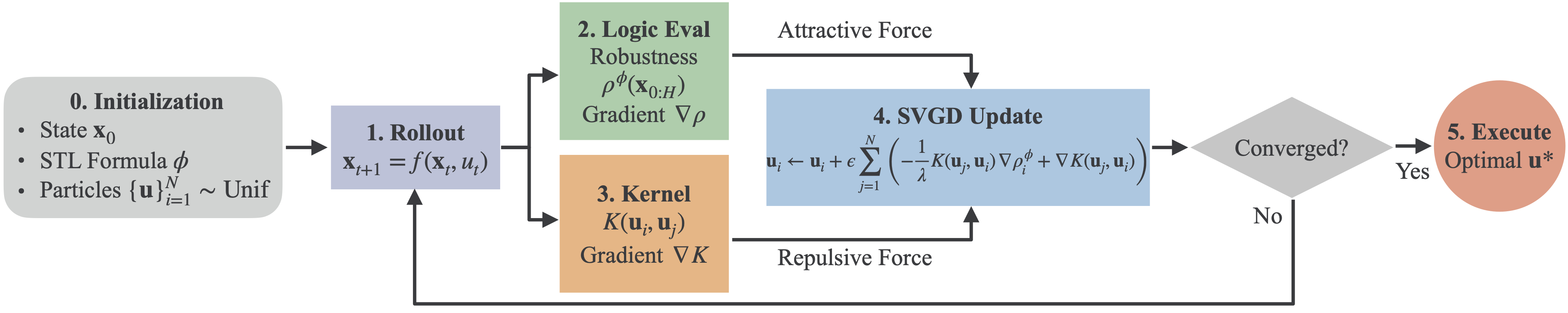}
    \caption{Overview of the STL-SVPIO Optimization Algorithm.}
    \label{fig:overview}
    \vspace{-15pt}
\end{figure*}

\section{Methodology}\label{sec:method}
\vspace{-1pt}
In this section, we formulate path integral control as a variational inference problem to optimize STL robustness. Because complex STL specifications create highly non-convex, multi-modal cost landscapes, we propose a non-parametric approach using Stein Variational Gradient Descent (SVGD). Our resulting algorithm, STL-SVPIO, iteratively transports a population of control particles toward an optimal distribution by balancing an attractive force driven by STL robustness gradients with a repulsive force that promotes diversity and prevents mode collapse.
We show a flowchart overview of our proposed algorithm in Figure \ref{fig:overview}. 

\subsection{Path Integral Control as Variational Inference}
Let $\mathbb{P}$ denote a prior distribution over control sequences with density $p(\mathbf{u})$.
Following the maximum entropy principle~\cite{jaynes1957information}, the resulting optimal control distribution takes the Gibbs form:
\begin{equation}\label{eq:optimal_def}
    \mathbb{P}^*: p^*(\mathbf{u})=\frac{1}{Z}\exp\left(-\frac{1}{\lambda}J(\mathbf{u})\right)p(\mathbf{u}),
\end{equation}
where $J(\mathbf{u})$ is the trajectory cost, $Z$ is the partition function that normalizes $p^*(\mathbf{u})$ into a valid distribution, and $\lambda$ is the temperature.
Since $p^*(\mathbf{u})$ is generally intractable to sample from, we introduce a variational proposal distribution $\mathbb{Q}$ with density $q(\mathbf{u})$.
The goal of the variational inference is to minimize the KL divergence between the proposal distribution and the target posterior defined as:
\begin{equation}\label{eq:kl_def}
\begin{aligned}
    D_{\text{KL}}(\mathbb{Q}\parallel\mathbb{P}^*) & = \mathbb{E}_{\mathbb{Q}} \left[\log\frac{q(\mathbf{u})}{p^*(\mathbf{u})}\right]\\
    & = \mathbb{E}_{\mathbb{Q}} \left[\log q(\mathbf{u}) -\log p^*(\mathbf{u})\right]. \\
\end{aligned}
\end{equation}
By substituting in the definition from Equation \eqref{eq:optimal_def}, we have:
\begin{equation}\label{eq:kl_next}
\begin{aligned}
D_{\mathrm{KL}}(\mathbb{Q}&\,\|\,\mathbb{P}^*)
= \mathbb{E}_{\mathbb{Q}}\!\left[
\log q(\mathbf{u})
+ \frac{1}{\lambda}J(\mathbf{u})
- \log p(\mathbf{u})
+ \log Z
\right] \\
&= \mathbb{E}_{\mathbb{Q}}\!\left[\frac{1}{\lambda}J(\mathbf{u})\right]
 + \mathbb{E}_{\mathbb{Q}}\!\left[\log q(\mathbf{u})-\log p(\mathbf{u})\right]
 + \log Z \\
&\propto
\mathbb{E}_{\mathbb{Q}}\!\left[\frac{1}{\lambda}J(\mathbf{u})\right]
 + \mathbb{E}_{\mathbb{Q}}\!\left[\log q(\mathbf{u})-\log p(\mathbf{u})\right].
\end{aligned}
\end{equation}
where $Z$ is a constant as it is independent of the variational distribution $q(\mathbf{u})$.

From here, approaches such as MPPI~\cite{williams_aggressive_2016} assume Gaussian distributions for the prior and proposal distributions, thereby reducing the problem to minimizing the expected trajectory cost and the quadratic control effort over the control sequence. However, this assumption is inadequate when optimizing STL robustness for tasks with complex logical structure, where the cost landscape is multi-modal.

We assume a uniform prior $\mathbb{P}: \mathbf{u}\sim \operatorname{Unif}(\mathbf{u}_\text{min},\mathbf{u}_\text{max})$, $p(\mathbf{u})=\frac{1}{V}\mathmybb{1}_{[\mathbf{u}_{\text{min}}, \mathbf{u}_{\text{max}}]}(\mathbf{u})$. Where $V$ is the volume of the control space and $\mathmybb{1}$ is the indicator function for inside the control bounds.
This choice reflects an unbiased prior over admissible control sequences, so all structure in the posterior arises solely from the trajectory cost.
The new target posterior $\mathbb{P}^*$ is now an exponentiated cost surface restricted by hard constraints:
\begin{equation}\label{eq:target}
    p^*(\mathbf{u}) = \frac{1}{ZV} \exp\left(-\frac{1}{\lambda} J(\mathbf{u})\right) \cdot \mathmybb{1}_{[\mathbf{u}_{\text{min}}, \mathbf{u}_{\text{max}}]}(\mathbf{u}).
\end{equation}
We obtain the free energy by multiplying Equation \eqref{eq:kl_next} by the temperature $\lambda$:
\begin{equation*}
    \mathcal{F} = \mathbb{E}_{\mathbb{Q}} [ J(\mathbf{u}) ] + \lambda \mathbb{E}_{\mathbb{Q}} [ \log q(\mathbf{u}) ] - \lambda \mathbb{E}_{\mathbb{Q}} [ \log p(\mathbf{u}) ].
\end{equation*}
Since our prior is now uniform, the logarithm is now constant: $\mathbb{E}_{\mathbb{{Q}}}\log p(\mathbf{u})=-\log(V)$ and irrelevant in the minimization.
Furthermore, the term $-\mathbb{E}_{\mathbb{Q}}\left[\log q(\mathbf{u})\right]$ is the Shannon entropy $\mathcal{H}(\mathbb{Q})$ of the proposal distribution. Hence the new objective now becomes minimizing:
\begin{equation}\label{eq:new_obj}
    \mathcal{F} = \mathbb{E}_{\mathbb{Q}} [ J(\mathbf{u}) ] - \lambda \mathcal{H}(\mathbb{Q}).
\end{equation}
In other words, the objective becomes minimizing the expected cost and maximizing the entropy of the proposal distribution. Note that the objectives in Equation \eqref{eq:new_obj} exactly mirror the formulation used in MaxEntRL~\cite{ziebart2010modeling}, which augments standard cost minimization with an entropy maximization term. Algorithms like Soft Actor-Critic (SAC)~\cite{haarnoja2018soft} have demonstrated practical success of the MaxEntRL principle. However, to effectively maintain this entropy and transport the non-parametric particle population across the complex STL landscape, we require a specialized inference mechanism. In the next section, we introduce Stein Variational Gradient Descent~\cite{liu_stein_2016} to solve this exact problem.

\subsection{Stein Variational Gradient Descent}\label{sec:background_svgd}

SVGD~\cite{liu_stein_2016} is a general-purpose iterative variational inference algorithm, which transports a set of particles to match the target distribution. SVGD first initializes particles $\{\mathbf{u}_i^0\}_{i=1}^N$, step size $\epsilon$, and kernel $K$. In each iteration, the particles are updated by $\mathbf{u}_i^{t+1}=\mathbf{u}_i^t+\epsilon\varphi^*(\mathbf{u}_i^t)$. The function $\varphi^*(\cdot)$ characterizes the optimal perturbation that maximally decreases the KL-divergence between the proposal distribution and the target posterior:
\begin{equation}\label{eq:stein_dir}
\begin{aligned}
\varphi&^*(\mathbf{u}_i)=\\&\frac{1}{N}\sum_{j=1}^N\bigg[ \underbrace{K(\mathbf{u}_j, \mathbf{u}_i) \nabla_{\mathbf{u}_j} \log p(\mathbf{u}_j)}_{\text{Attractive Force}} 
   + \underbrace{\nabla_{\mathbf{u}_j} K(\mathbf{u}_j, \mathbf{u}_i)}_{\text{Repulsive Force}} \bigg].
\end{aligned}
\end{equation}
Intuitively, the two terms in the summation represent the attraction force towards the target distribution and the repulsion force generated by the kernel to prevent particles from collapsing, respectively. Note that the kernel appears in both terms, enabling ``information sharing" between particles to determine the best optimization direction. Without the kernel multiplication, the attraction force reduces to individual gradient descent. Next, we will show how SVGD is used to perform path-integral optimization

\subsection{STL-guided Stein Variational Path Integral Optimization}
Following SVGD, the proposal distribution $\mathbb{Q}$ is now represented non-parametrically by a population of particles $\{\mathbf{u}\}_{i=0}^N$.
Instead of taking a softmax-weighted average of rollouts, as in Importance Sampling, we use SVGD updates to push particles in $\mathbb{Q}$ towards matching the target posterior $p^*(\mathbf{u})$.
SVGD relies on the derivative of the target's log probability to minimize the expected cost. By taking the log of our target posterior in Equation \eqref{eq:target}:
\begin{equation}\label{eq:svgd_score_deriv}
    \begin{aligned}
        \log p^*(\mathbf{u}) &= -\frac{1}{\lambda} J(\mathbf{u}) - \log Z - \log V, \\
        \nabla_{\mathbf{u}} \log p^*(\mathbf{u})&= \nabla_{\mathbf{u}} \left( -\frac{1}{\lambda} J(\mathbf{u}) - \log Z - \log V\right) \\
        & = -\frac{1}{\lambda} \nabla_{\mathbf{u}} J(\mathbf{u}), ~~~~~\mathbf{u}\in[\mathbf{u}_{\min},\mathbf{u}_{\max}].
    \end{aligned}
\end{equation}
With a uniform prior, the score function of the target posterior depends solely on the gradient of the trajectory cost, simplifying the SVGD update and avoiding additional prior-induced forces.
For the derivation for transforming the entropy of the proposal in Equation \eqref{eq:new_obj} into the kernel repulsion force, we refer the readers to~\cite{liu_stein_2016}. The general idea is to increase the volume of the proposal distribution by maximizing the divergence of a vector field (the trace of the Jacobian of the proposal distribution). When restricting that vector field to a reproducing kernel Hilbert space, the divergence of the vector field is exactly the gradient of the chosen kernel. In summary, the optimal perturbation direction $\varphi^*(\mathbf{u})$ for a particle $\mathbf{u}_i$ is given by the following SVGD update:
\begin{equation}\label{eq:svgd_update}
\begin{aligned}\varphi&^*(\mathbf{u}_i)=\\& \frac{1}{N} \sum_{j=1}^N \Bigg[ -\frac{1}{\lambda}K(\mathbf{u}_j, \mathbf{u}_i) \nabla_{\mathbf{u}_j} J(\mathbf{u}_j) + \nabla_{\mathbf{u}_j} K(\mathbf{u}_j, \mathbf{u}_i) \Bigg], \\
\mathbf{u}_i &\leftarrow \mathbf{u}_i + \epsilon \varphi^*(\mathbf{u}_i),
\end{aligned}
\end{equation}
where $K(\cdot, \cdot)$ is a positive definite kernel.

We choose negative STL robustness evaluated on the state trajectory $\mathbf{x}_{i,0:H}$ induced by a particle $\mathbf{u}_i$ as the cost function:
\begin{equation}
    J(\mathbf{u}_i) = -\rho^\phi(\mathbf{x}_{i,0:H}).
\end{equation}
Following~\cite{lambert_stein_2021}, we use a smooth RBF kernel:
\begin{equation}
    K_{\text{RBF}}(\mathbf{u}_i,\mathbf{u}_j)=\exp\left\{\frac{ -\parallel\mathbf{u}_i-\mathbf{u}_j \parallel_2^2}{h}\right\},
\end{equation}
where $h$ is the median heuristic on the population of all particles: $h=\text{median}(\{\mathbf{u}_i\})^2/\log(N-1)$. For a specific STL formula $\phi$, the SVGD update step in our algorithm is as follows:
\begin{equation}\label{eq:stl_svgd_update}
\begin{aligned}
\varphi^*(\mathbf{u}_i) &= \frac{1}{N} \sum_{j=1}^N \bigg[\frac{1}{\lambda} {K_{\text{RBF}}(\mathbf{u}_j, \mathbf{u}_i) \nabla_{\mathbf{u}_j} \rho^\phi\left(\mathbf{x}_{j,0:H}\right)}\\
&~~~~~~~~~~~~~~~~~~~~~~~~ + \nabla_{\mathbf{u}_j} K_{\text{RBF}}(\mathbf{u}_j, \mathbf{u}_i) \bigg], \\
\mathbf{u}_i &\leftarrow \mathbf{u}_i + \epsilon \varphi^*(\mathbf{u}_i).
\end{aligned}
\end{equation}
Note that we assume that the dynamics and integration method chosen to generate the state trajectory are differentiable. Recent advances in computation frameworks such as JAX~\cite{jax2018github} make automatic differentiation through physics simulations tractable. 

\subsection{Algorithm Summary}

\begin{algorithm}
\caption{STL-SVPIO}
\label{alg:stl_svpio}
\begin{algorithmic}[1]
\Require
Initial state $\mathbf{x}_0$,
simulation function $\texttt{sim}()$,
STL specification $\phi$,
horizon $H$,
number of particles $N$,
optimization iterations $M$,
step size $\epsilon$,
temperature $\lambda$.

\State \textbf{Initialize:} Sample $\{\mathbf{u}\}_{i=1}^N \sim \operatorname{Unif}(\mathbf{u}_{\min}, \mathbf{u}_{\max})$


\For{iter $m \gets 1$ to $M$}
    \State $\forall_{ij}$ Compute $K(\mathbf{u}_i,\mathbf{u}_j)$, $\nabla_{\mathbf{u}_i}K(\mathbf{u}_j, \mathbf{u}_i)$
    \For{$i\gets 1$ to $N$}\Comment{\textbf{In parallel}}
    
    \State Rollout all particles $\mathbf{x}^i_{0:H} \gets \texttt{sim}(\mathbf{x}^i_0, \mathbf{u}_i)$
    \State Compute attraction $s_i=-\frac{1}{\lambda}\nabla_{\mathbf{u}_i} \rho^\phi(\mathbf{x}^i_{0:H})$
    
    \State $\Delta\mathbf{u}_i \leftarrow \frac{1}{N}\sum_{j=1}^N \left[K(\mathbf{u}_j,\mathbf{u}_i)s_j + \nabla_{\mathbf{u}_j}K(\mathbf{u}_j, \mathbf{u}_i)\right]$
    
    \State $\mathbf{u}_{i}\leftarrow \mathbf{u}_i + \epsilon \Delta \mathbf{u}_i$
    \EndFor
\EndFor

\Return $\argmax_{\mathbf{u}\in\{\mathbf{u}_{i=1}^N\}}\rho_i^\phi$
\end{algorithmic}
\end{algorithm}

The STL-SVPIO algorithm (Algorithm \ref{alg:stl_svpio}) proceeds as follows:
First, sample $N$ particles as control sequences that spans the entire horizon $H$ from a uniform distribution $\mathbf{u}^i_{0:H} \sim \operatorname{Unif}(\mathbf{u}_{\min}, \mathbf{u}_{\max})$.
In each iteration, for each particle in parallel, rollout the dynamics $f$ to get trajectories $\mathbf{x}_{0:H}^{i}$ with the \texttt{sim()} function.
Next, compute the pairwise kernel value and gradient for each particle.
Then compute STL robustness $\rho^\phi$ and its gradient $\nabla_{\mathbf{u}} \rho^\phi$ as the attraction force.
Lastly, transport all particles using SVGD.
At the end of the iterations, we select the particle with the highest STL robustness as the final solution.
\section{Experiments}\label{sec:exp}

In this section, we evaluate STL-SVPIO through three sets of experiments. 
First, we consider a simple reach-avoid task with 2D point-mass dynamics to demonstrate the wide range of baselines considered.
Second, we demonstrate the effectiveness and efficiency of STL-SVPIO on more complex scenarios in both single- and multi-agent settings. For the single-agent case, we focus on long-horizon planning in cluttered environments. For the multi-agent case, we evaluate the methods' ability to enforce task-completion ordering, achieve synchronized goal satisfaction, and resolve inter-agent conflicts. Finally, we show that STL-SVPIO generalizes to highly agile motion planning tasks with nonlinear dynamics, without requiring any modification to the algorithm.

We now introduce the baselines used for comparison. These baselines are chosen to highlight different aspects of our proposed method.
\begin{itemize}
  \item MPPI+Heuristics ~\cite{williams_aggressive_2016} is a sampling-based baseline with a handcrafted stage and terminal costs based on Euclidean distances to obstacles and the goal.
  \item SVMPC~\cite{lambert_stein_2021} is an ablation for STL-SVPIO, isolating the effect of exact STL gradients versus finite-difference approximations.
  \item DPI~\cite{halder_trajectory_2025} is a path integral baseline that was designed to solve STL specifications.
  \item STLCG++~\cite{leung2023backpropagation,kapoor_stlcg_2025} is another ablation, isolating the effect of STL-SVPIO's particle-based SVGD method versus direct gradient descent.
  \item MILP~\cite{belta_formal_2019,cardona_flexible_2023} is an STL solver with linear encoding.
\end{itemize}
In all experiments, each task is defined with an STL specification, and the robustness value of a trajectory is calculated following the recursive definition in Section \ref{sec:background_stl}. 
To evaluate the methods, the main metric we focus on is the robustness value each method's final solution achieves. The higher the robustness, the better. A task is satisfied when the robustness value is greater than zero. We use the \texttt{stljax} library from STLCG++~\cite{kapoor_stlcg_2025} for all robustness evaluation. For sampling-based methods, we evaluate performance over multiple different random seeds in Section \ref{sec:larg_scale_exp}. All experiments are conducted on NVDIA RTX A6000 GPUs.

\subsection{Single Agent Reach Avoid}
\label{sec:simple_exp}
We first evaluate STL-SVPIO on a simple single-agent reach–avoid task using a 2D point mass model. The system is a double integrator with acceleration inputs.
\begin{figure}[h]
  \centering
  \includegraphics[width=0.48\textwidth]{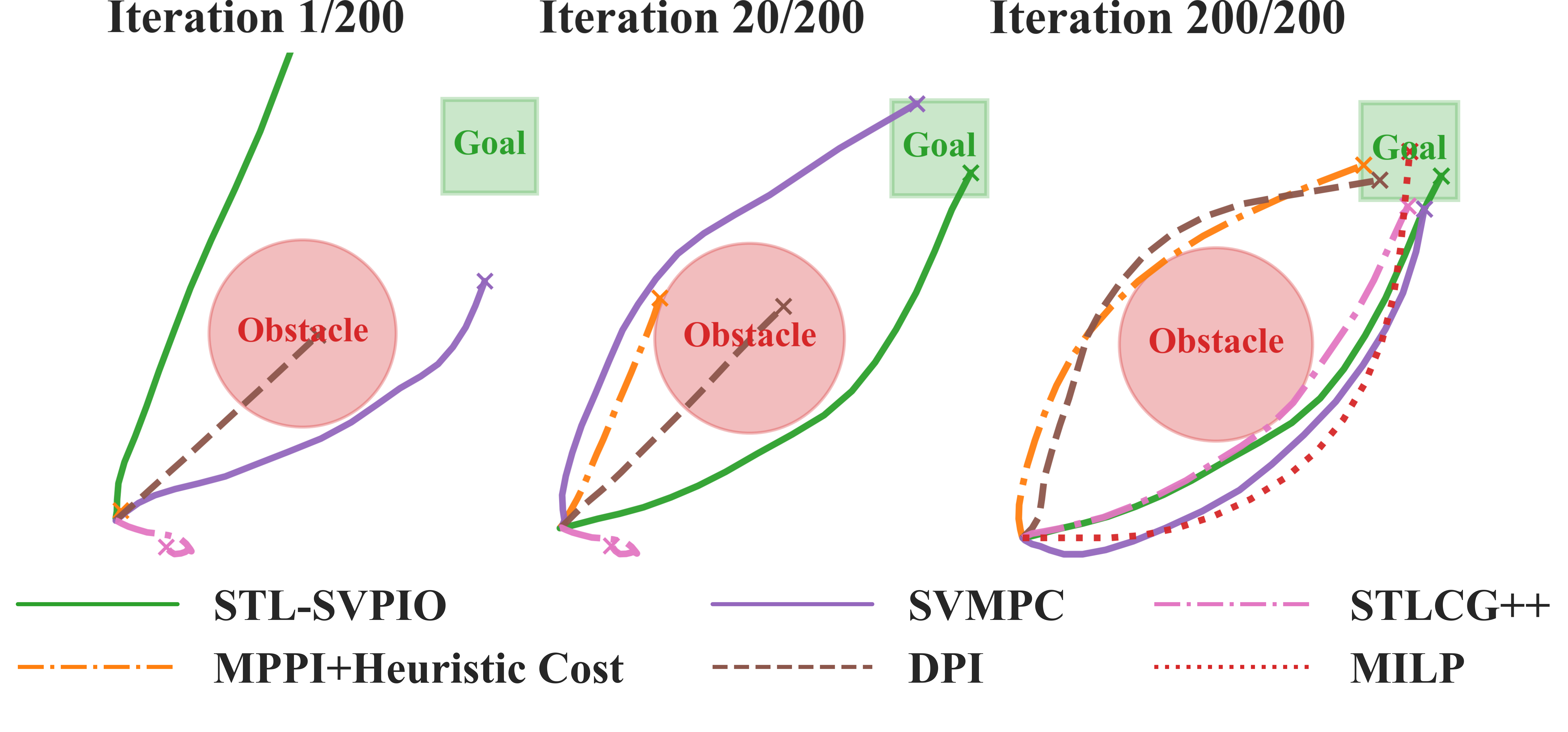}
  \caption{Selected trajectories of each method through iterations.}
  \label{fig:toy_comparison}
\end{figure}
The task requires the agent to always avoid a circular obstacle, and eventually reach the square goal zone within the planning horizon $H$. The corresponding STL specification is:
\begin{equation}\label{eq:toy_formula}
    \phi_{\text{ra}}:=\always_{[0,H]}\mu_{\text{avoid}}(\mathbf{x}_t) \wedge \eventually_{[0,H]}\mu_{\text{in\_box}}(\mathbf{x}_t),
\end{equation}
with predicates:
\begin{equation*}
\begin{aligned}
    \mu_{\text{avoid}}(\mathbf{x}_t):=&\parallel \mathbf{x}_t - c_{\text{obs}} \parallel_2 > r_{\text{obs}}, \\
    \mu_{\text{in\_box}}(\mathbf{x}_t):=&\parallel \mathbf{x}_t - c_{\text{goal}} \parallel_\infty \leq \frac{1}{2}h_{\text{goal}},
\end{aligned}
\end{equation*}
where $c_{\text{obs}}$ and $r_{\text{obs}}$ denote the obstacle center and radius, while $c_{\text{goal}}$ and $h_{\text{goal}}$ define the center and the side length of the goal zone.

\begin{table}[h]
    \centering
    \setlength{\tabcolsep}{3pt}
    \begin{tabular}{|c|c|c|c|c|}
    \hline
         Method & \# Samples & \# Iterations & Runtime (ms) & Robustness \\
    \hline
         STL-SVPIO & 10   & 20  & 19.152 & 0.108 \\
         MPPI      & 10   & 200 & 187.125 &0.005\\
         SVMPC     & 64   & 200 & 178.389 &-0.098\\
         DPI       & 1024 & 200 & 194.492 &0.179\\
         STLCG++   & N/A    & 200 & 926.704 &-0.068\\
         MILP      & N/A  & N/A & 324.578 & 0.495\\
    \hline
    \end{tabular}
    \caption{Reach avoid task parameters and robustness}
    \label{tab:toy_results}
    \vspace{-15pt}
\end{table}

Figure \ref{fig:toy_comparison} shows each algorithm's selected trajectories during optimization, and Table \ref{tab:toy_results} compares parameters used, runtime, and final robustness values produced by each method on the single-agent reach–avoid task. SVMPC and STLCG++ only find solutions that violate the specification by not arriving at the goal zone at the final step. All other methods find satisfying solutions.
Our method achieves positive robustness with the lowest number of samples and iterations. While MPPI+Heuristics performs well with handcrafted costs, it does not naturally generalize beyond this task. In this simple setting, the finite-difference gradients used by SVMPC are sufficient to reach near-feasible solutions but remain less efficient than those of STL-SVPIO. DPI already exhibits high sample usage in a simple task. STLCG++ shows slow convergence. MILP achieves the best satisfaction margin while maintaining a reasonable solve time. However, we show in subsequent experiments that it does not scale with more complicated STL formulae. Although most methods succeed on this reach–avoid task, the efficiency gap becomes more pronounced in the more complex single- and multi-agent experiments.

\subsection{More Complex 2D Point Mass Tasks}
\label{sec:larg_scale_exp}
\begin{figure*}
    \centering
    \includegraphics[width=0.86\linewidth]{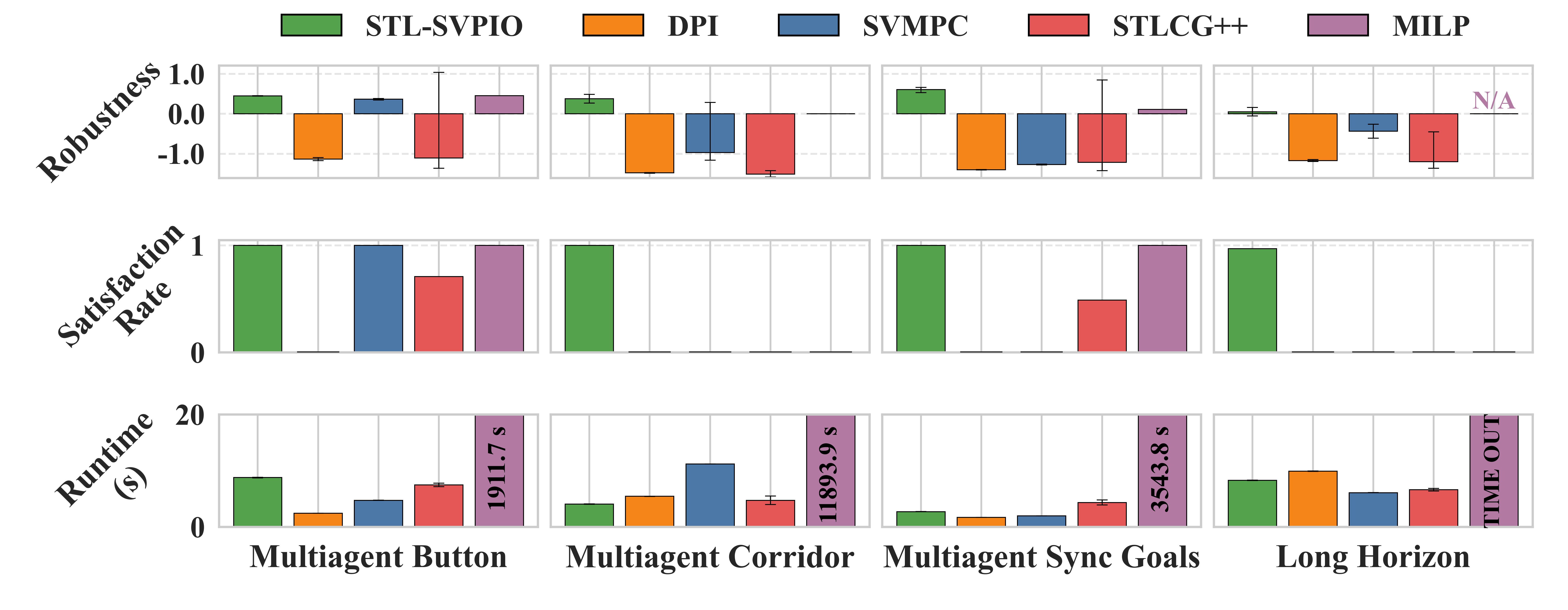}
    \vspace{-15pt}
    \caption{Aggregated average robustness and satisfaction rate of methods. All algorithms besides MILP were tested across 100 different random sampling seeds. Each solution with positive robustness counts as a satisfaction. Multiagent Button corresponds to the scenario in Section \ref{sec:exp_button}, Multiagent Corridor corresponds to the scenario in Section \ref{sec:exp_corridor}, Multiagent Sync Goals corresponds to the scenario in Section \ref{sec:exp_sync}, and Long Horizon corresponds to the scenario in Section \ref{sec:exp_long_horizon}. Gurobi was given a 10-hour time limit to solve the MILP.}
    \label{fig:barchart}
    \vspace{-12pt}
\end{figure*}
We next evaluate STL-SVPIO on a set of more challenging single- and multi-agent tasks. All agents are the same 2D point-mass as in Section \ref{sec:simple_exp}. Before introducing each task, we define a set of atomic predicates to construct the corresponding STL formulae. Agents are indexed by $i\in\{0,\ldots,n\}$, obstacles by $o$, and zones by $z$.
Agent collision radii are denoted by $r_{\text{col}}$. Obstacle centers and radii are denoted by $c$ and $r$. We define the following predicates:
\begin{itemize}
\item Obstacle avoidance for agent $i$ and obstacle $o$ at time $t$:
    \begin{equation*}
        \mu_{\text{avoid}}(i,o,t) := \|\mathbf{x}_i(t) - c_o^{\text{obs}} \|_2 > r_o^{\text{obs}},
    \end{equation*}
\item Agent $i$ inside circular zone $z$ at time $t$:
    \begin{equation*}
        \mu_{\text{in}}(i,z,t) :=\|\mathbf{x}_i(t)-c_z^{\text{zone}}\|_2 \leq r_z^{\text{zone}},
    \end{equation*}
\item Agent $i$ inside box zone $z$ with side lengths $h_z^x$ and $h_z^y$ with center $(c_x^{\text{zone}}, c_y^{\text{zone}})$:
    \begin{equation*}
    \begin{aligned}
        \mu_{\text{in\_box}}(i,z,t) := \min\{&h_z^x-|(\mathbf{x}_i(t))_x-c_x^{\text{zone}}|,\\
        &h_z^y-|(\mathbf{x}_i(t))_y-c_y^{\text{zone}}|\} > 0,
    \end{aligned}
    \end{equation*}
\item Pairwise collision avoidance between agents $i$ and $j$:
    \begin{equation*}
        \mu_{\text{col}}(i,j,t) := \|\mathbf{x}_i(t)-\mathbf{x}_j(t)\|_2 \geq 2r_{\text{col}}.
    \end{equation*}
\end{itemize}
We design four experiments:
\subsubsection{Long Horizon Planning in Cluttered Environments}\label{sec:exp_long_horizon}
\begin{figure}[h]
    \centering
    \includegraphics[width=0.5\linewidth]{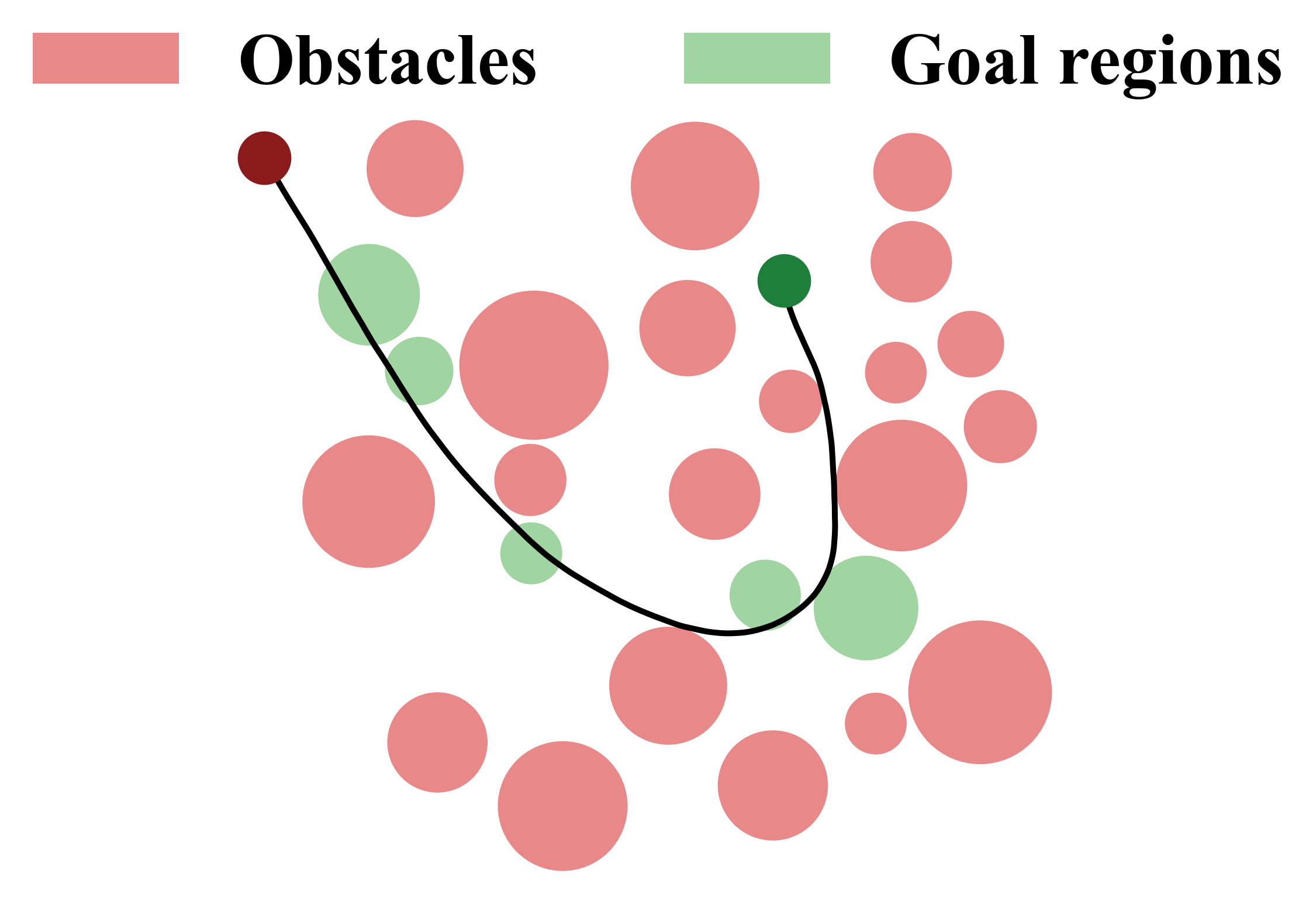}
    \caption{Long horizon trajectory synthesized by STL-SVPIO through a cluttered obstacle field. Dark green and red markers indicate the start and final states, respectively.}
    \label{fig:single_long_horizon}
\end{figure}
In this task (Figure \ref{fig:single_long_horizon}), a single agent must traverse a cluttered environment and reach all assigned goals within a 600-step horizon. The STL specification is:
\begin{equation*}
    \phi_{\text{lh}}:=\Big(\bigwedge_{o} \always_{[0,H]} \mu_{\text{avoid}}(1, o, t)\Big) \land \Big(\bigwedge_{z}\eventually_{[0, H]}\mu_{\text{in}}(1,z,t)\Big)
\end{equation*}
This scenario evaluates the ability to find feasible trajectories through narrow passages over long horizons.

\subsubsection{Enforcing Task Completion Order}\label{sec:exp_button}
\begin{figure}[h]
    \centering
    \includegraphics[width=1.0\linewidth]{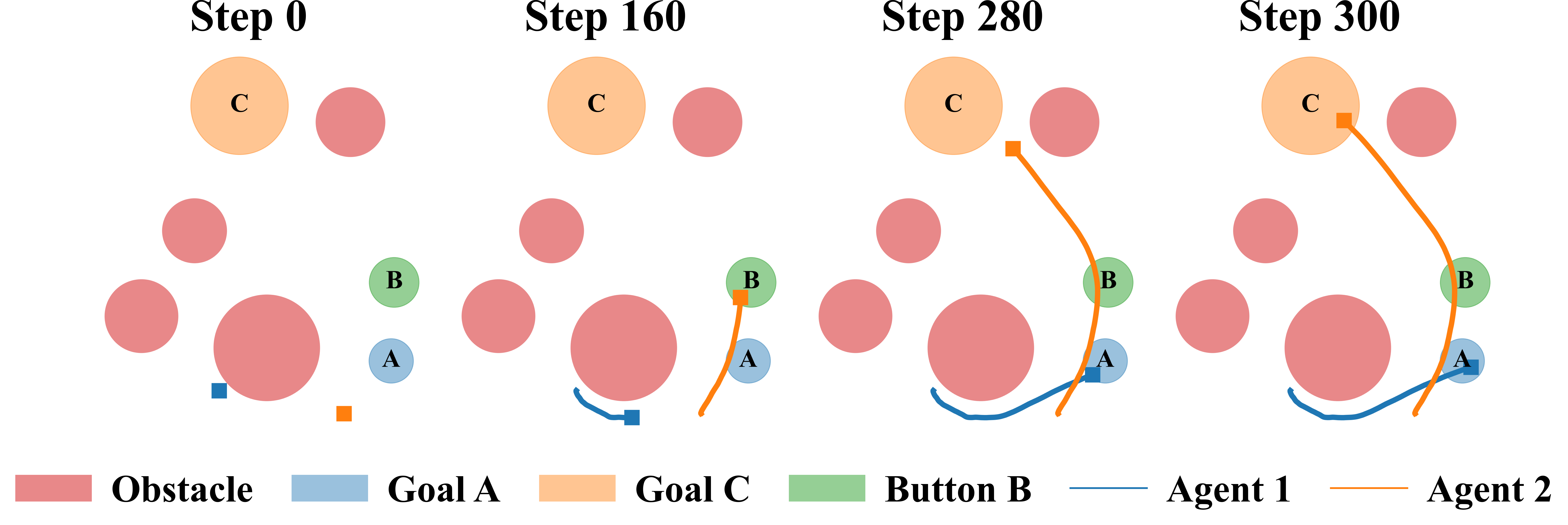}
    \caption{STL-SVPIO synthesizes a multi-agent trajectory that satisfies strict ordering constraints between task events.}
    \label{fig:button_solution}
\end{figure}

In this multi-agent task (Figure \ref{fig:button_solution}), agent $1$ must reach goal $A$ but may only do so after agent $2$ activates button $B$. Agent $2$ must eventually reach goal $C$. Both agents must avoid all obstacles at all times.
Gated entry is specified by:
\begin{equation*}
    \phi_{\text{gated\_entry}} := \neg\mu_{\text{in}}(2,C,t)\mathbf{U}\mu_{\text{in}}(1, B, t).
\end{equation*}
The full STL specification for this task is:
\begin{equation*}
\begin{aligned}
    \phi_{\text{co}} := \Big(\bigwedge_o \always_{[0,H]}\mu_{\text{avoid}}(1,o,t)\Big) \land \Big(\bigwedge_o \always_{[0,H]}\mu_{\text{avoid}}(2,o,t)\Big)\\
    \land \Big(\eventually_{[0,H]}\mu_{\text{in}}(1,A,t)\Big) \land \Big(\eventually_{[0,H]}\mu_{\text{in}}(2,C,t)\Big) \land \phi_{\text{gated\_entry}}.
\end{aligned}
\end{equation*}

This task tests strict temporal ordering constraints in a coupled multi-agent setting. 

\subsubsection{Synchronized Task Completion}\label{sec:exp_sync}
\begin{figure}[h]
    \centering
    \includegraphics[width=0.85\linewidth]{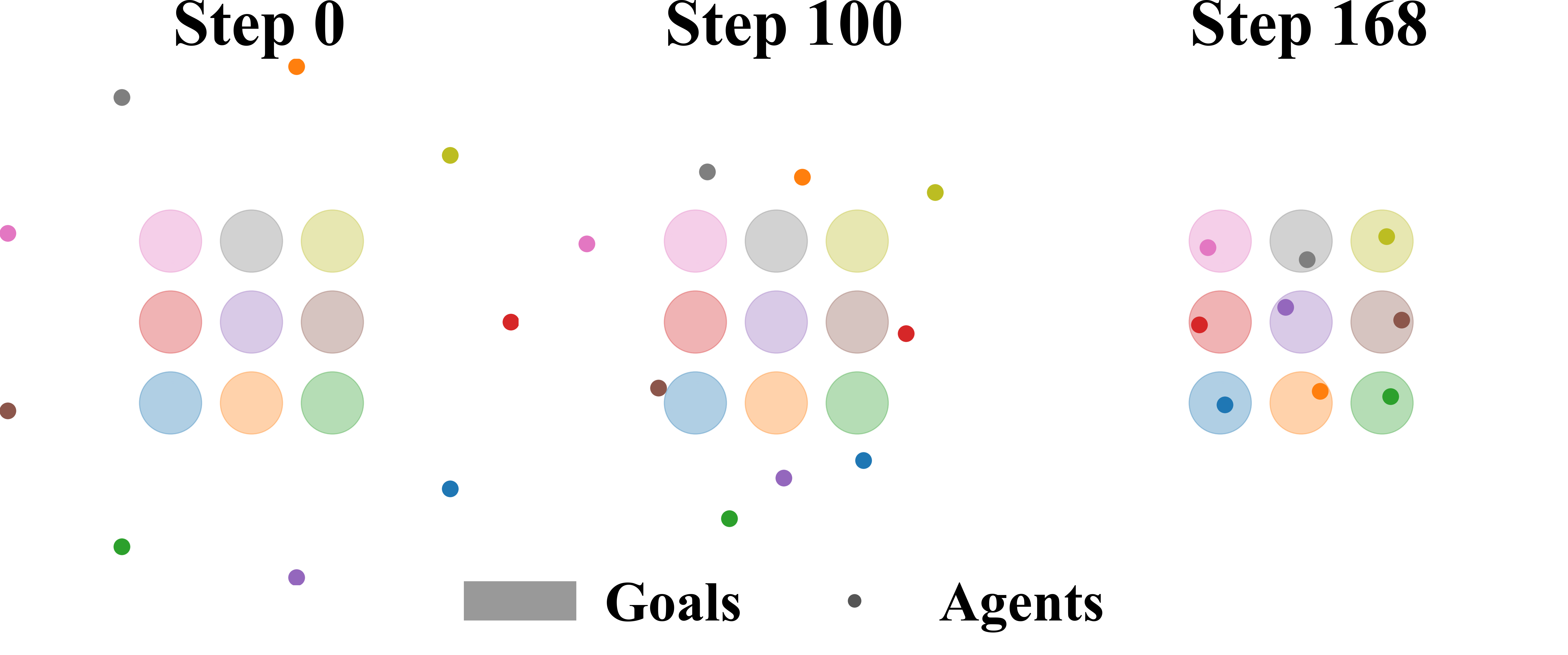}
    \caption{A solution generated by STL-SVPIO in which all agents reach their assigned goals within a shared, unspecified time window.}
    \label{fig:sync_goal_solution}
\end{figure}

In this task (Figure \ref{fig:sync_goal_solution}), all agents must reach their respective goals within a shared but unspecified time window of width $2\delta$, while avoiding collisions. The STL specification of this task is:
\begin{equation*}
\begin{aligned}
    \phi_{\text{sg}} := \Big(\bigwedge_{i<j}\always_{[0,H]}\mu_{\text{col}}(i,j,t) \Big) \land \eventually_{[0,H-2\delta]}\Big( \bigwedge_{i}\eventually_{[0,2\delta]}\mu_{\text{in}}(i,i,t) \Big).
\end{aligned}
\end{equation*}

This scenario evaluates coordination in high-dimensional joint spaces under tight temporal coupling.

\subsubsection{Conflict Resolution}\label{sec:exp_corridor}
\begin{figure}[h]
    \centering
    \includegraphics[width=1.0\linewidth]{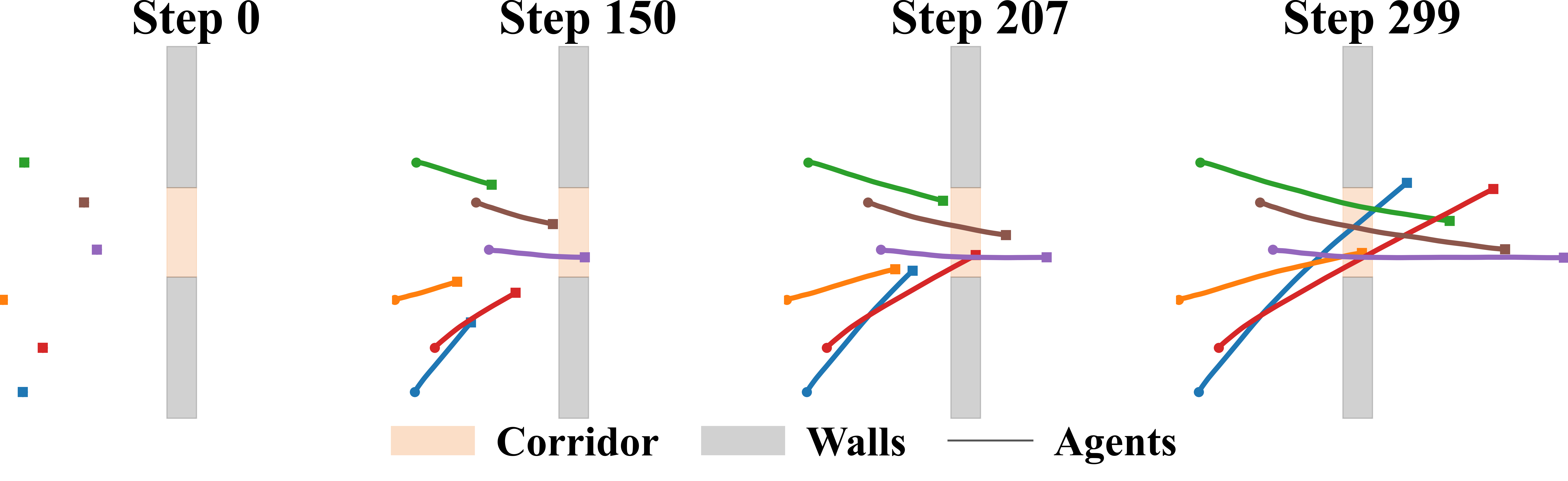}
    \caption{STL-SVPIO plans collision-free trajectories that enforce sequential corridor occupancy among multiple agents.}
    \label{fig:corridor_solution}
\end{figure}

In this task (Figure \ref{fig:corridor_solution}), agents must traverse a wall at $x=0$ through a narrow corridor, allowing only one agent inside the corridor at any time. Let
\begin{equation*}
    \phi_{x^+}(i,t):=(\mathbf{x}_i(t))_x \geq 0,
\end{equation*}
and define exclusive corridor occupancy:
\begin{equation*}
    \phi_{\text{occ}}(i,j,z,t):=\neg\big( \mu_{\text{in\_box}}(i,z,t) \land \mu_{\text{in\_box}}(j,z,t) \big).
\end{equation*}
We index the rectangular boxes that define the walls with $w$, and the corridor rectangular box as $z_{\text{cor}}$.
The STL specification used for this task is:
\begin{equation*}
\begin{aligned}
    \phi_{\text{cr}} = &\Big(\bigwedge_{i<j}\always_{[0,H]}\mu_{\text{col}}(i,j,t) \Big) \land \Big( \bigwedge_i\bigwedge_w \always_{[0,H]}\neg\mu_{\text{in\_box}}(i, w, t) \Big) \\
    &\land \Big(\bigwedge_i\eventually_{[0,H]}\mu_{\text{in\_box}}(i,z_{\text{cor}},t)\Big) \land \Big( \bigwedge_i\eventually_{[0,H]}\mu_{x^+}(i,t) \Big) \\
    &\land \Big(\bigwedge_{i<j}\always_{[0,H]}\phi_{\text{occ}}(i,j,z_{\text{cor}},t)\Big)
\end{aligned}
\end{equation*}
This task enforces both collision avoidance and implicit queuing behavior in a tightly constrained joint space.

\textbf{Baseline Comparison Results:} We compare STL-SVPIO against DPI, SVMPC, STLCG++, and MILP using average robustness, satisfaction rate, and runtime (Figure \ref{fig:barchart}). MPPI with handcrafted costs is omitted, as designing dense stage and terminal costs for these intertwined temporal constraints is impractical. All stochastic methods are evaluated over 100 random seeds, while MILP is solved once per task. For each algorithm and scenario, we perform an extensive hyperparameter sweep over 100 to 400 configurations, depending on the method, and report the best-performing results to ensure a fair and best-effort comparison.

MILP demonstrates exactness when solutions are found, but suffers from severe scalability limitations. With a 10-hour time limit, MILP succeeds only in the Button (\ref{sec:exp_button}) and Sync Goals (\ref{sec:exp_sync}) tasks and exhibits runtimes orders of magnitude longer. In the Corridor task, linear approximations of Euclidean distance predicates lead to solutions that satisfy the MILP but violate the original STL specification, resulting in zero satisfaction. In the Long-Horizon task, MILP times out due to problem size and geometric complexity.

Compared to STLCG++, STL-SVPIO benefits from maintaining a particle population. Although both methods use exact STL gradients, STLCG++ shows high variance and fails in the Corridor (\ref{sec:exp_corridor}) and Long-Horizon (\ref{sec:exp_long_horizon}) tasks due to high sensitivity to initialization and local minima.

Among sampling-based methods, STL-SVPIO is the only approach that succeeds across all scenarios. DPI struggles due to its uni-modal proposal distribution and reliance on terminal costs, making successful samples extremely rare in highly non-convex STL landscapes, even with 1024–2048 samples per iteration, compared to 10 particles for STL-SVPIO. SVMPC performs well in simpler tasks with 64 particles, but finite-difference gradients become unreliable in harder scenarios and incur high effective sampling costs (8x number of particles used) when estimating the gradient. Overall, these results demonstrate that combining exact STL gradients with particle-based variational updates enables robust and scalable planning in complex multi-agent settings.

\subsection{Nonlinear Dynamic Motion Planning}
We next demonstrate that STL-SVPIO generalizes beyond 2D point-mass systems to motion planning problems with highly nonlinear dynamics. We use MuJoCo MJX~\cite{todorov2012mujoco}, a JAX~\cite{jax2018github} based physics engine that supports GPU parallel simulation and automatic differentiation under appropriate solver and integrator settings. In our implementation, MJX is used as the parallel forward dynamics module (Algorithm 1, line 7), enabling end-to-end differentiability through the simulation.

\subsubsection{Franka Panda Goal Reach}
\begin{figure}[h]
    \centering
    \includegraphics[width=0.25\linewidth]{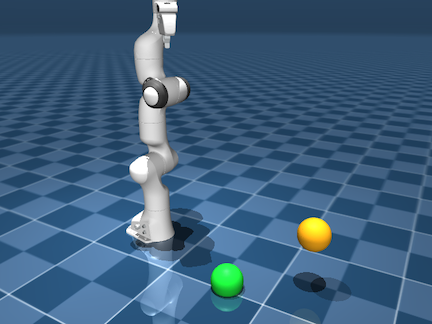}%
    \includegraphics[width=0.25\linewidth]{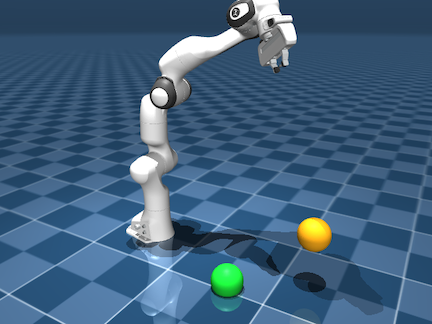}%
    \includegraphics[width=0.25\linewidth]{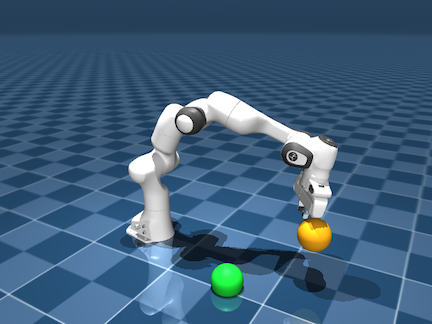}%
    \includegraphics[width=0.25\linewidth]{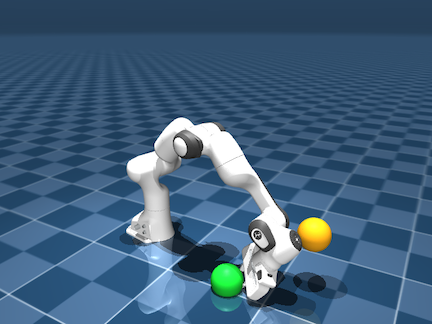}%
    \caption{Rollout snapshots of a torque-controlled Franka Panda trajectory synthesized by STL-SVPIO. The end effector first reaches the yellow goal and subsequently reaches the green goal.}
    \label{fig:panda_solution}
    \vspace{-10pt}
\end{figure}

Figure \ref{fig:panda_solution} shows a 7-DoF Franka Panda manipulation task in which the end effector must reach multiple goal regions. Let $\mathbf{x}_{ee}(t)$ denote the end-effector position, and let each goal $g$ be defined by center $c_g$ radius $r_g$.
The reach predicate is:
\begin{equation*}
    \mu_{\text{reach}}(g, t):=\|\mathbf{x}_{ee}(t)-c_g\|_2 \leq r_g,
\end{equation*}
and the STL specification is:
\begin{equation*}
    \phi_{\text{panda}}=\bigwedge_g\eventually_{[0, H]}\mu_{\text{reach}}(g, t).
\end{equation*}

STL-SVPIO requires no algorithmic modification beyond hyperparameter tuning. Control particles are defined directly in the joint torque space, and no inverse kinematics are used. Using only 10 particles and 300 iterations, STL-SVPIO successfully synthesizes a torque-level control sequence that reaches all goals in 634.85 seconds, with a final robustness value of 0.029, satisfying the STL specification. The solution's trajectory are shown in Figure \ref{fig:panda_solution} as snapshots.

\subsubsection{Half Cheetah Back Flip}
\begin{figure}[ht!]
    \centering
    \includegraphics[trim={240px 250px 250px 390px},clip,width=0.125\linewidth]{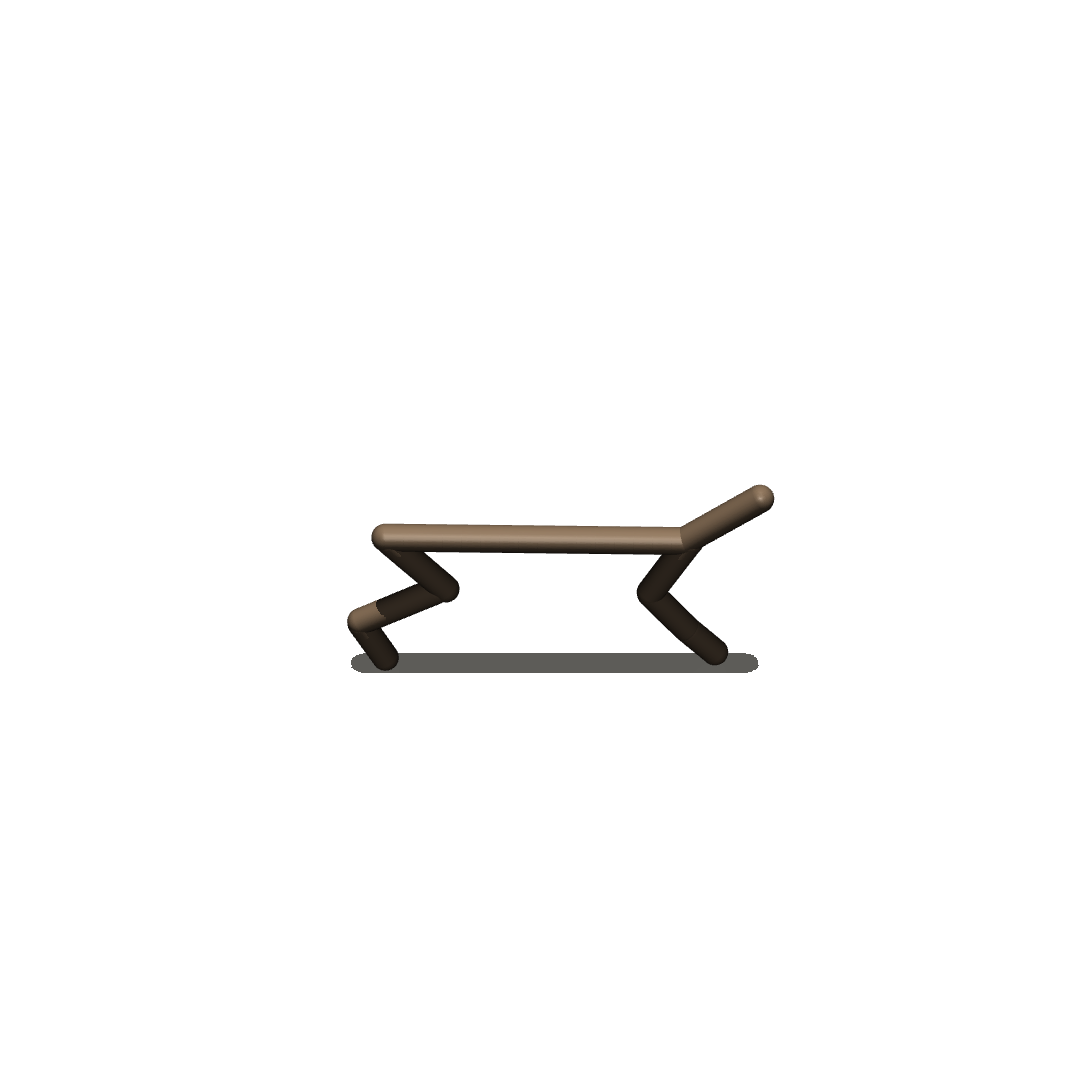}%
    \includegraphics[trim={240px 250px 250px 390px},clip,width=0.125\linewidth]{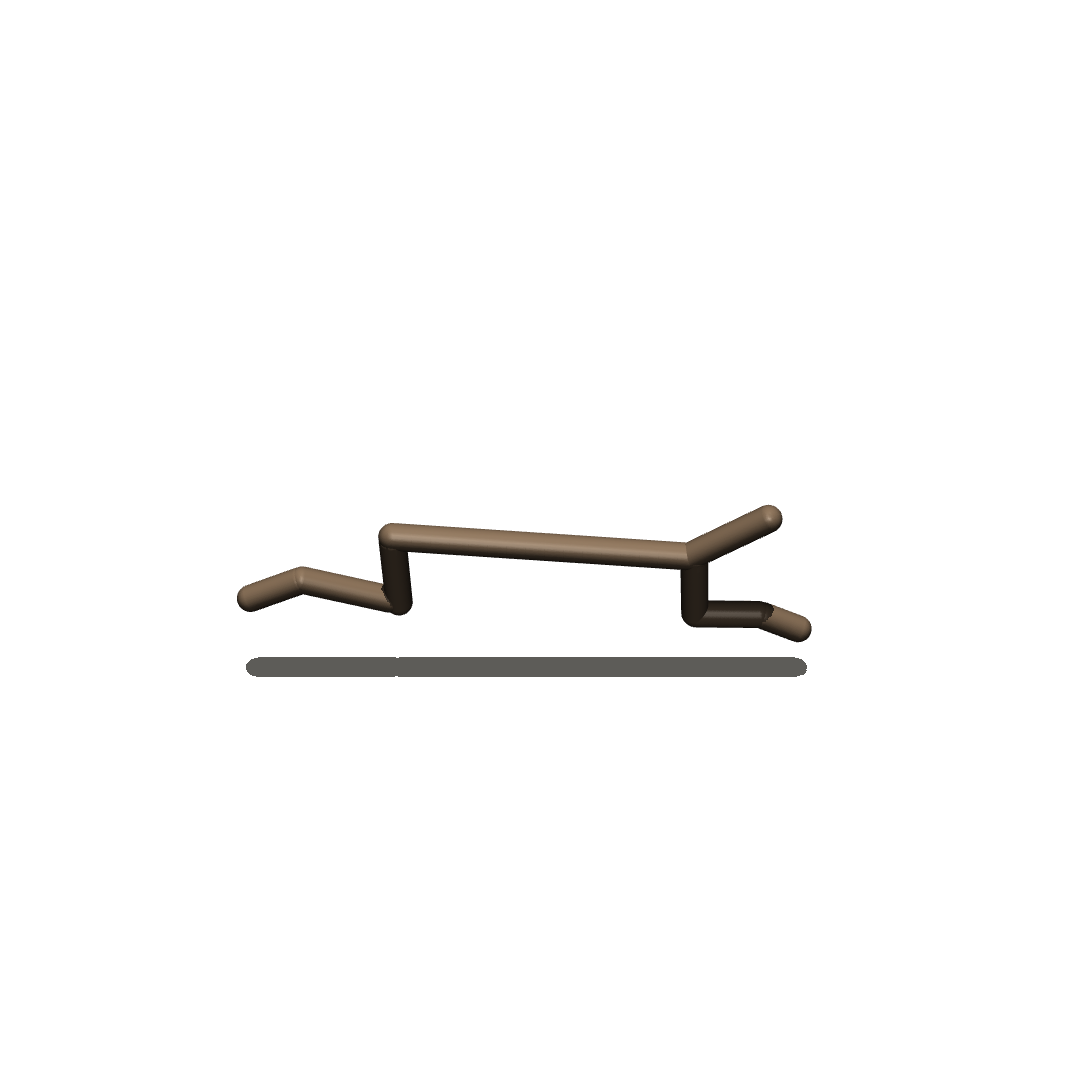}%
    \includegraphics[trim={240px 250px 250px 390px},clip,width=0.125\linewidth]{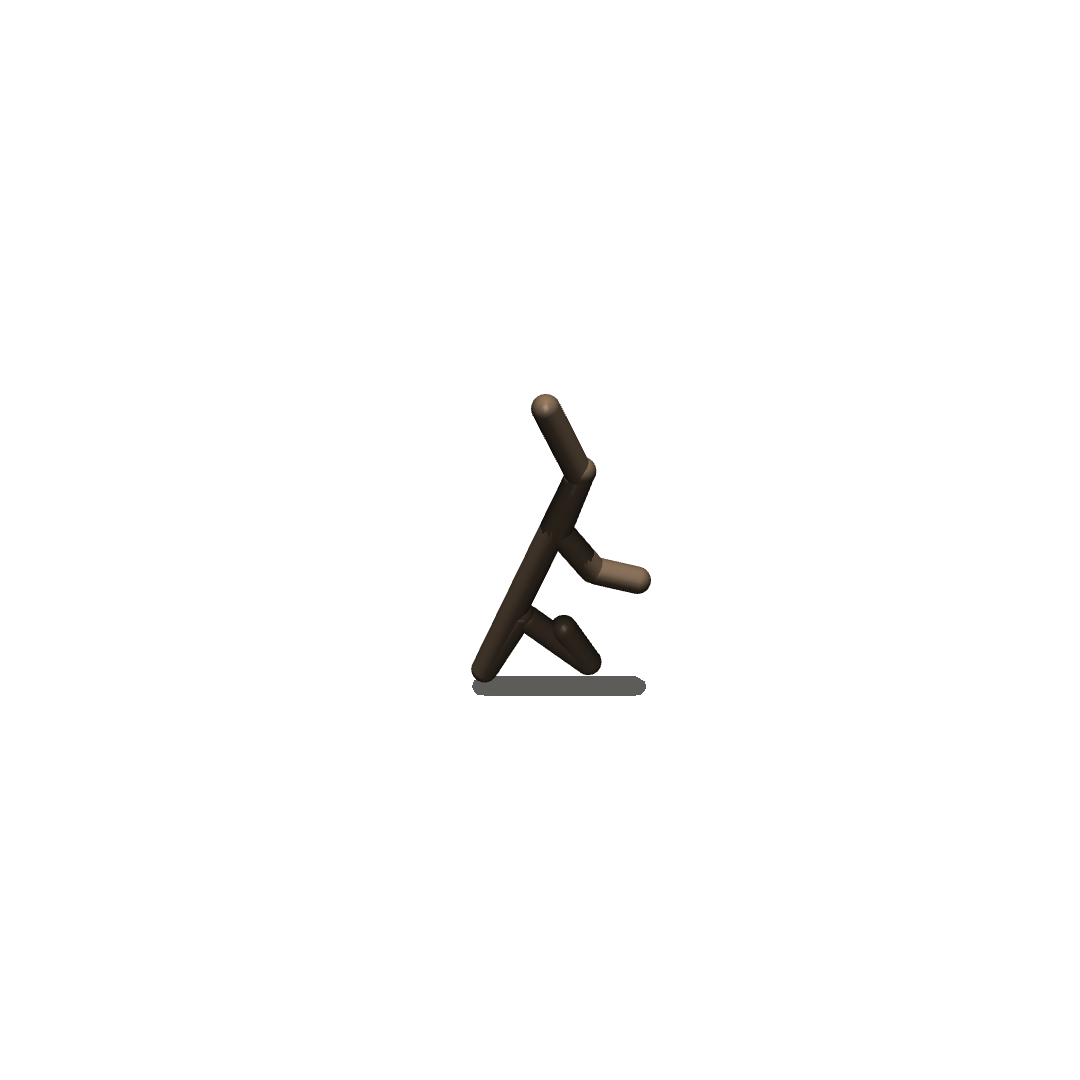}%
    \includegraphics[trim={240px 250px 250px 390px},clip,width=0.125\linewidth]{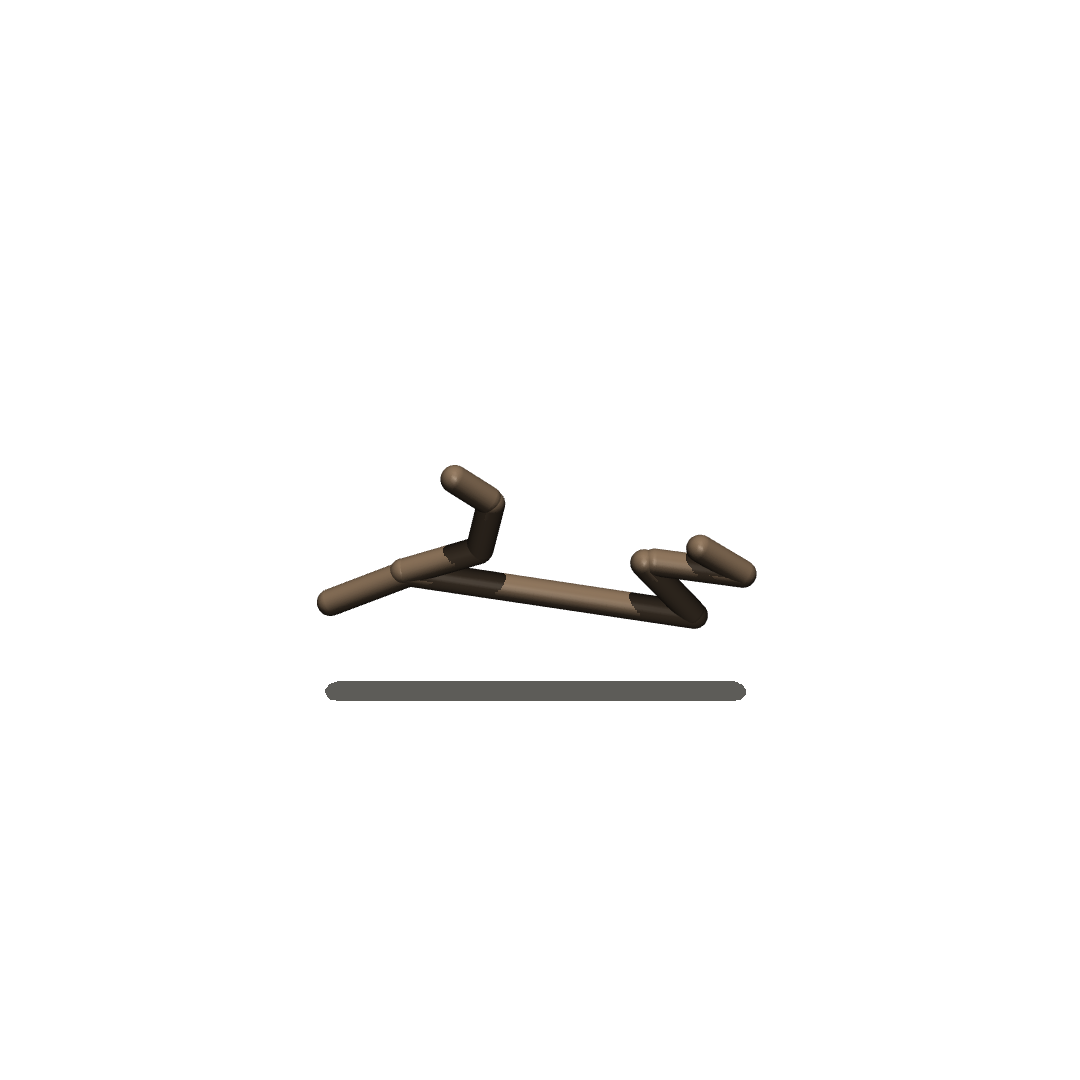}%
    \includegraphics[trim={240px 250px 250px 390px},clip,width=0.125\linewidth]{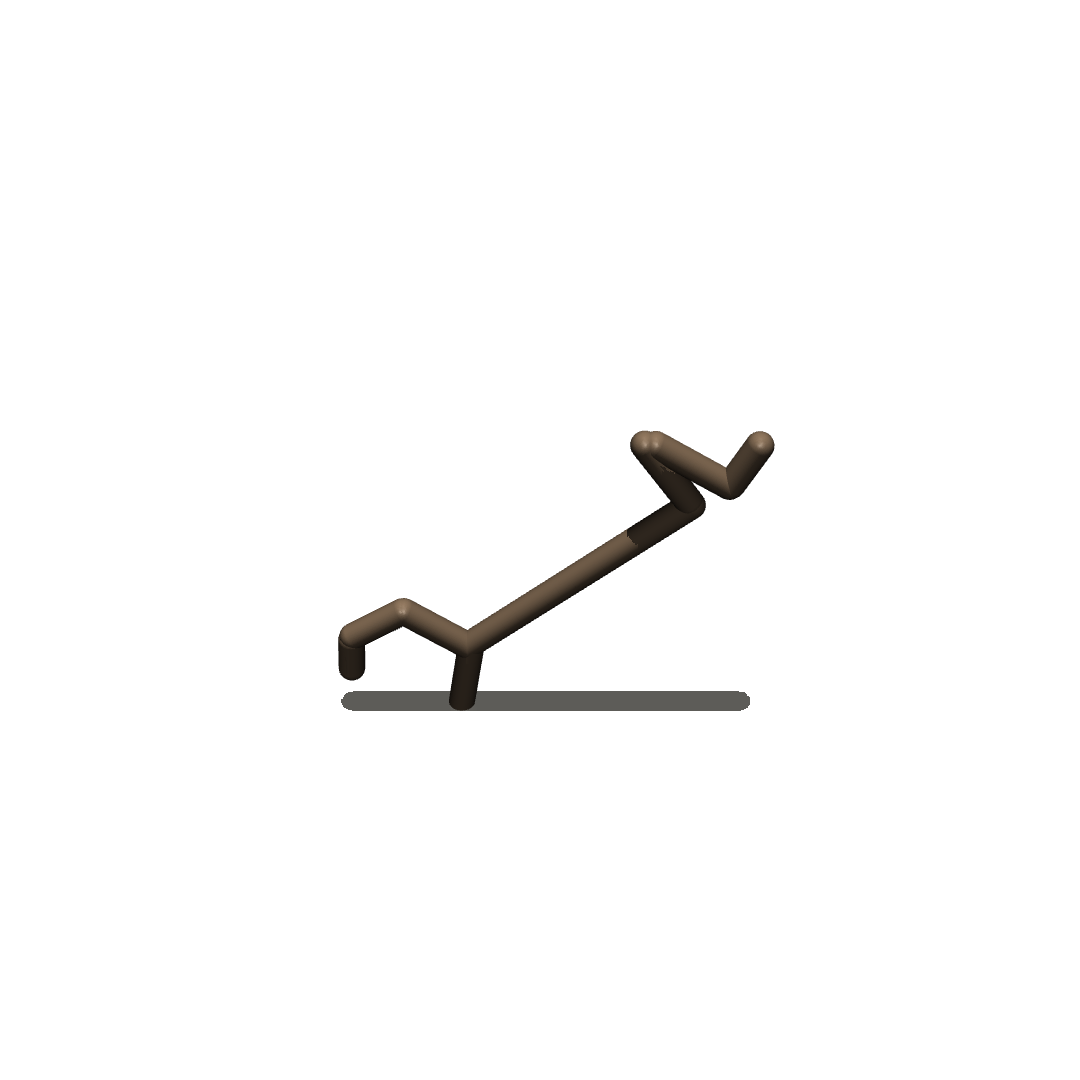}%
    \includegraphics[trim={240px 250px 250px 390px},clip,width=0.125\linewidth]{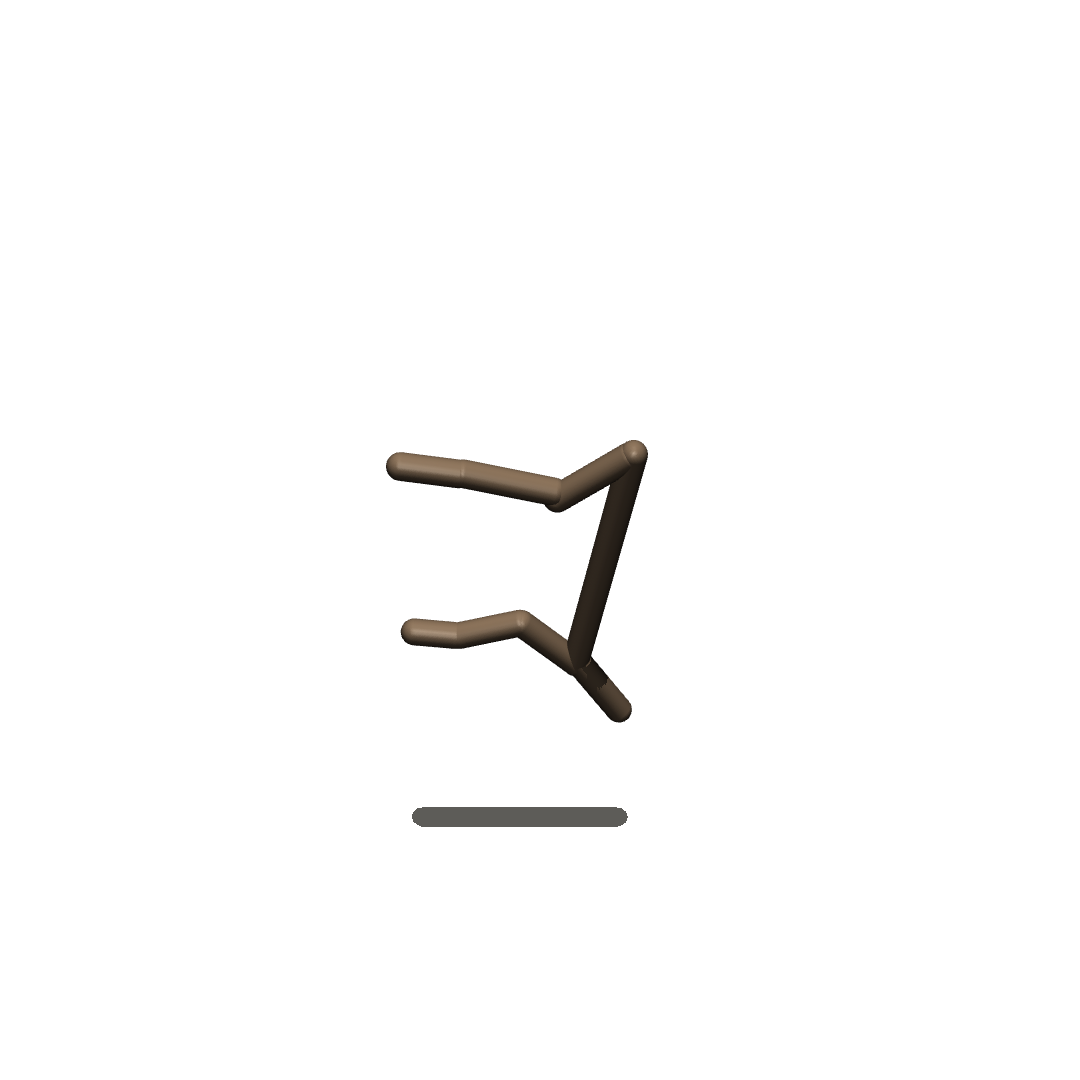}%
    \includegraphics[trim={240px 250px 250px 390px},clip,width=0.125\linewidth]{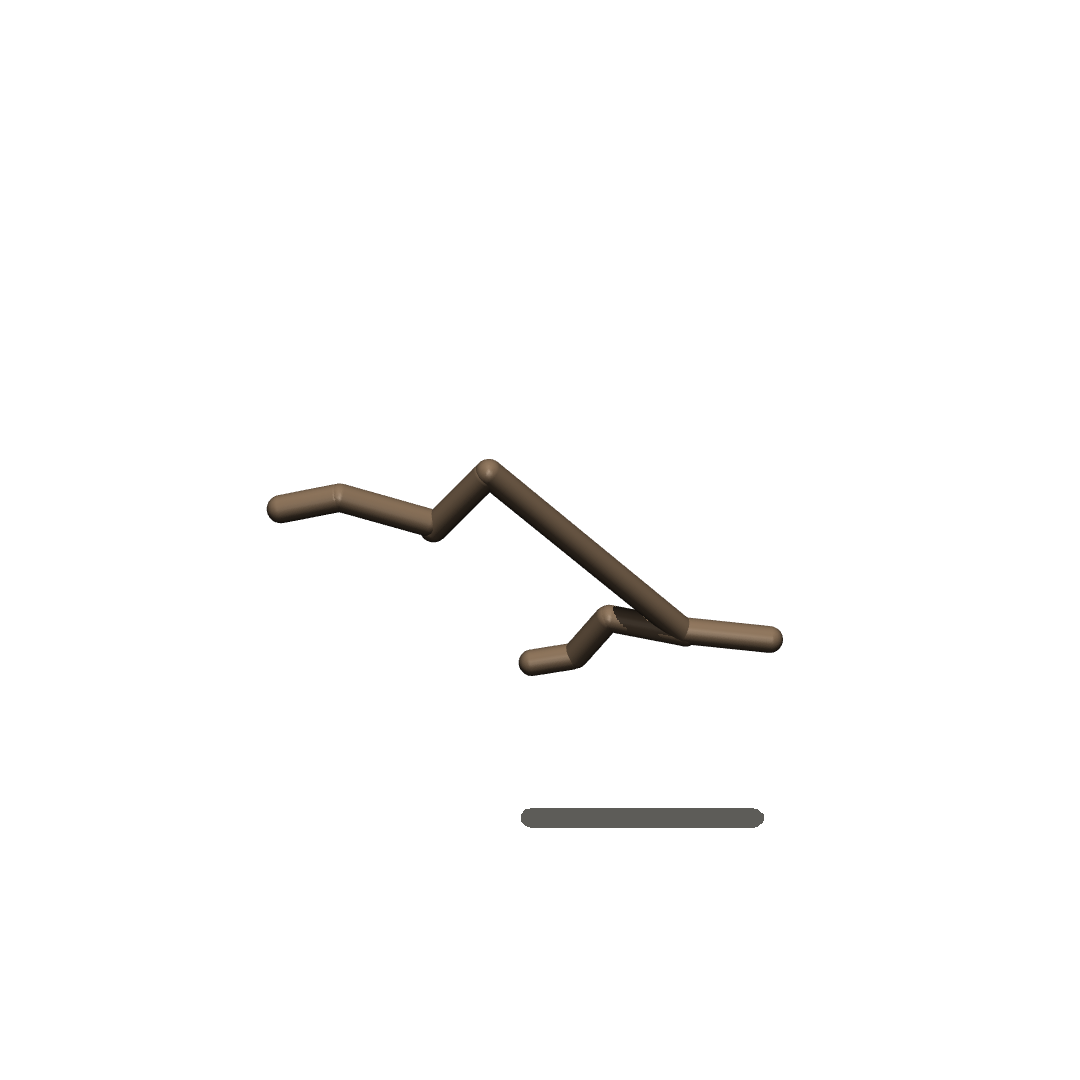}%
    \includegraphics[trim={240px 250px 250px 390px},clip,width=0.125\linewidth]{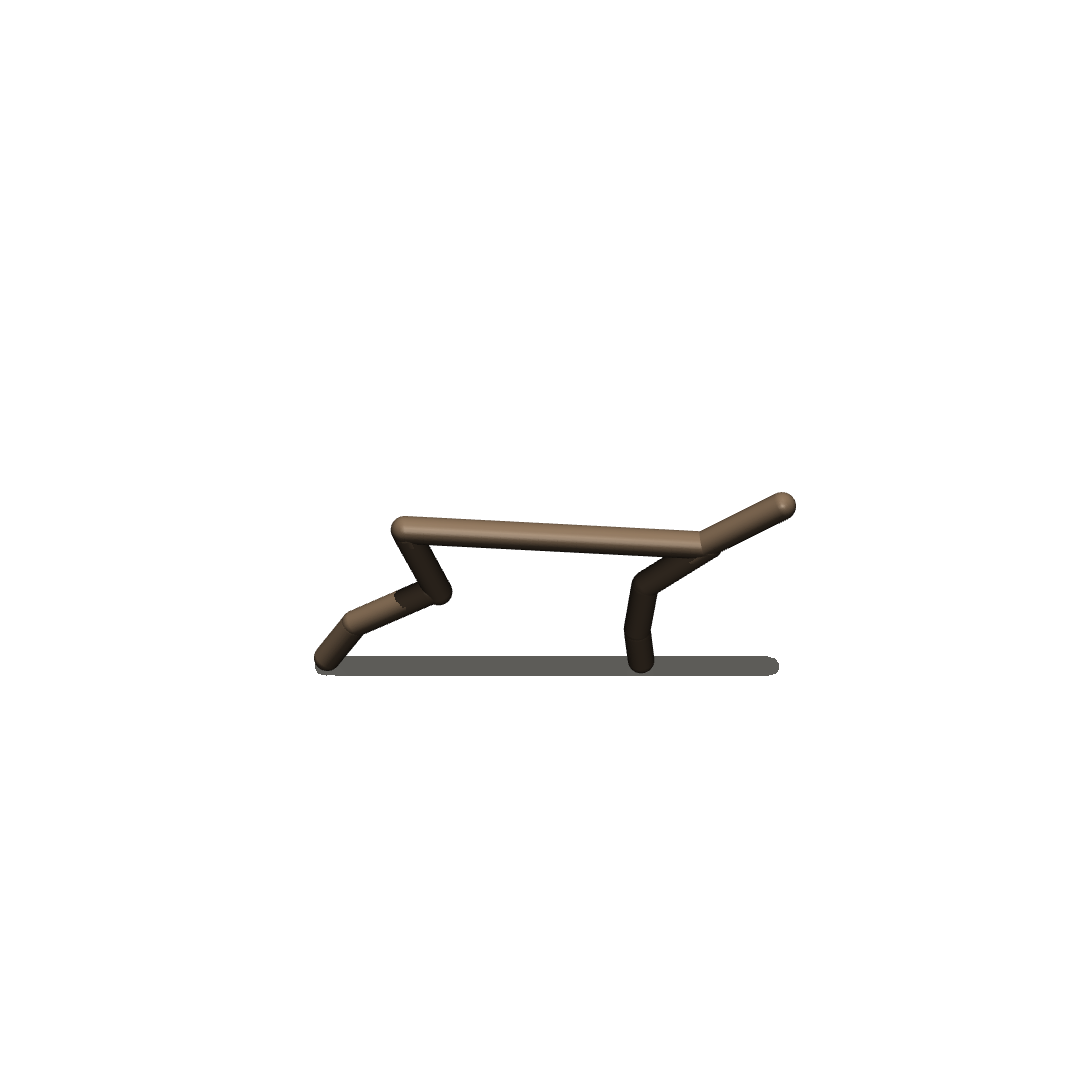}%
    \caption{Rollout snapshots of a Half-Cheetah back flip trajectory synthesized by STL-SVPIO.}
    \label{fig:cheetah_solution}
    \vspace{-5pt}
\end{figure}

Highly agile motions such as back flips are difficult to specify and optimize using standard reward functions. In reinforcement learning, such behaviors often rely on extensive reward shaping, human feedback~\cite{christiano2017deep}, or open-ended skill discovery~\cite{meier2022open}. In contrast, we specify the desired physical effects of a back flip directly using STL.

Let $z(t)$ denote the torso height, $\theta(t)$ the torso pitch angle, $\omega(t)$ the pitch rate.
We define predicates $\mu_{z}(t) := z(t) > z_{\min}$ for torso height, $\mu_{\omega}(t) :=  |\omega(t)| < \omega_{\max}$ for safe pitch rate, and $\mu_{\text{flip}}(t):=-(\theta(t)-\theta(0)) > \theta_{\text{target}}-\epsilon_\theta$ for flip completion.
The full STL specification is:
\begin{equation*}
    \phi_{\text{back\_flip}}=\always_{[0, H]}\mu_{z}(t) \land  \always_{[0,H]}\mu_{\omega}(t) \land \eventually_{[H-\delta, H]}\mu_{\text{flip}}(t).
\end{equation*}
Figure \ref{fig:cheetah_solution} shows the resulting rollout. STL-VPIO obtained a solution using 10 particles and 300 iterations in 593.22 seconds, with a final robustness value of 0.168, satisfying the specification. Despite the highly nonlinear dynamics and discontinuous contact events, STL-SVPIO successfully discovers an agile backflip motion without any task-specific modifications to the algorithm.
\section{Conclusion and Limitation}
We presented STL-SVPIO, a variational framework that synthesizes continuous control trajectories from STL specifications. By integrating differentiable STL robustness with SVGD and differentiable physics, STL-SVPIO efficiently navigates long-horizon, non-convex landscapes. It outperforms prior baselines in robustness and scalability, generalizing seamlessly to complex nonlinear systems.

However, our reliance on auto-differentiable physics means high-fidelity simulations incur steep memory and runtime costs. Additionally, contact dynamics introduce discontinuities that require careful hyperparameter tuning. Future work will extend STL-SVPIO to real-world tasks and motion planning for physical platforms, including quadrupeds, humanoids, and quadrotors.


\section*{ACKNOWLEDGMENT}
Gemini and ChatGPT were used for fixing grammatical errors and improving clarity on the non-technical sections of the manuscript. Codex was used to assist in implementing experimental code. All technical content, experimental design, results, and conclusions were developed, verified, and carefully vetted by the authors.

\bibliographystyle{IEEEtran}
\bibliography{references}

\end{document}